# Sliced Recurrent Neural Networks

**Zeping Yu**
School of Eclectronic Information and
Electrical Engineering
Shanghai Jiao Tong University
zepingyu@foxmail.com

**Gongshen Liu***
School of Eclectronic Information and
Electrical Engineering
Shanghai Jiao Tong University
lgshen@sjtu.edu.cn

**Abstract**

Recurrent neural networks have achieved great success in many NLP tasks. However, they have difficulty in parallelization because of the recurrent structure, so it takes much time to train RNNs. In this paper, we introduce sliced recurrent neural networks (SRNNs), which could be parallelized by slicing the sequences into many subsequences. SRNNs have the ability to obtain high-level information through multiple layers with few extra parameters. We prove that the standard RNN is a special case of the SRNN when we use linear activation functions. Without changing the recurrent units, SRNNs are 136 times as fast as standard RNNs and could be even faster when we train longer sequences. Experiments on six large-scale sentiment analysis datasets show that SRNNs achieve better performance than standard RNNs.

## 1 Introduction

Recurrent neural networks (RNNs) have been widely used in many NLP tasks, including machine translation (Cho et al., 2014; Bahdanau et al., 2015; Luong et al., 2015; Bradbury and Socher, 2016), question answering (Xiong et al., 2016; Chen et al., 2017), image caption (Xu et al., 2015; Karpathy and Li, 2015), and document classification (Tang et al., 2015; Yang et al., 2016; Zhou et al., 2016). RNNs have the ability to obtain the order information of the input sequences. The two most popular recurrent units are long short-term memory (LSTM) (Hochreiter and Schmidhuber, 1997) and gated recurrent unit (GRU) (Cho et al., 2014), both of which could store previous memory in hidden states and use a gating mechanism to determine how much previous memory should be combined with the current input. Unfortunately, because of the recurrent structure, RNNs cannot be computed in parallel. Therefore, training RNNs takes much time, which limits academic research and industrial applications.

To solve this problem, several scholars try to use convolutional neural networks (CNNs) (Lecun et al., 1998) instead of RNNs in the field of NLP (Kim, 2014; Kalchbrenner et al., 2014; Gehring et al., 2017). However, CNNs may not obtain the order information of the sequences, which is very important in NLP tasks.

Some scholars tried to increase the speed of RNNs by improving the recurrent units and they have achieved good results. Quasi-recurrent neural networks (QRNNs) (Bradbury et al., 2017) got up to 16 times faster speeds by combining CNNs with RNNs. Lei et al. (2017) proposed the simple recurrent unit (SRU), which is 5-10 times faster than LSTM. Similarly, strongly-typed recurrent neural networks (T-RNN) (Balduzzi and Ghifary, 2016) and minimal gated unit (MGU) (Zhou et al., 2016) are also methods that could change the recurrent units.

Although RNNs have achieved faster speeds in these researches with the recurrent units improved, the recurrent structure among the entire sequence remains unchanged. As we still have to wait for the output of previous step, the bottleneck of RNNs still exists. In this paper, we introduce sliced recurrent neural networks (SRNNs), which are substantially faster than standard RNNs without changing the recurrent units. We prove that when we use linear activation functions, the standard RNN is a special case of the SRNN, and the SRNN has the ability to obtain high-level information of the sequences.

* Corresponding author

In order to compare our model with the standard RNN, we choose GRU as the recurrent unit. Other recurrent units could also be used in our structure, because we improve the overall RNN structure among the whole sequence rather than just changing the recurrent units. We complete experiments on six large-scale datasets and SRNNs perform better than standard RNNs on all the datasets. We open source our implementation in Keras (François et al, 2015).[1]

## 2 Model Structure

### 2.1 Gated Recurrent Unit

The GRU (Bahdanau et al., 2014) has the reset gate *r* and the update gate *z*. The reset gate decides how much of the previous memory is combined with the new input, and the update gate determines how much of the previous memory is retained.

$$r_t = \sigma(W_r x_t + U_r h_{t-1} + b_r) \tag{1}$$

$$z_t = \sigma(W_z x_t + U_z h_{t-1} + b_z) \tag{2}$$

where x is the input and h is the hidden state.

$$\tilde{h}_t = \tanh(W_h x_t + U_h (r_t \circ h_{t-1}) + b_h) \tag{3}$$

The candidate hidden state $\tilde{h}_t$ is controlled by the reset gate. When the reset gate is 0, the previous memory is ignored.

$$h_t = z_t \circ h_{t-1} + (1 - z_t) \circ \tilde{h}_t \tag{4}$$

When the update gate is 1, the hidden state could copy the previous memory to the current moment and ignore the current input.

### 2.2 The Standard RNN Structure

The standard RNN structure is shown in Figure 1, where A denotes the recurrent units.

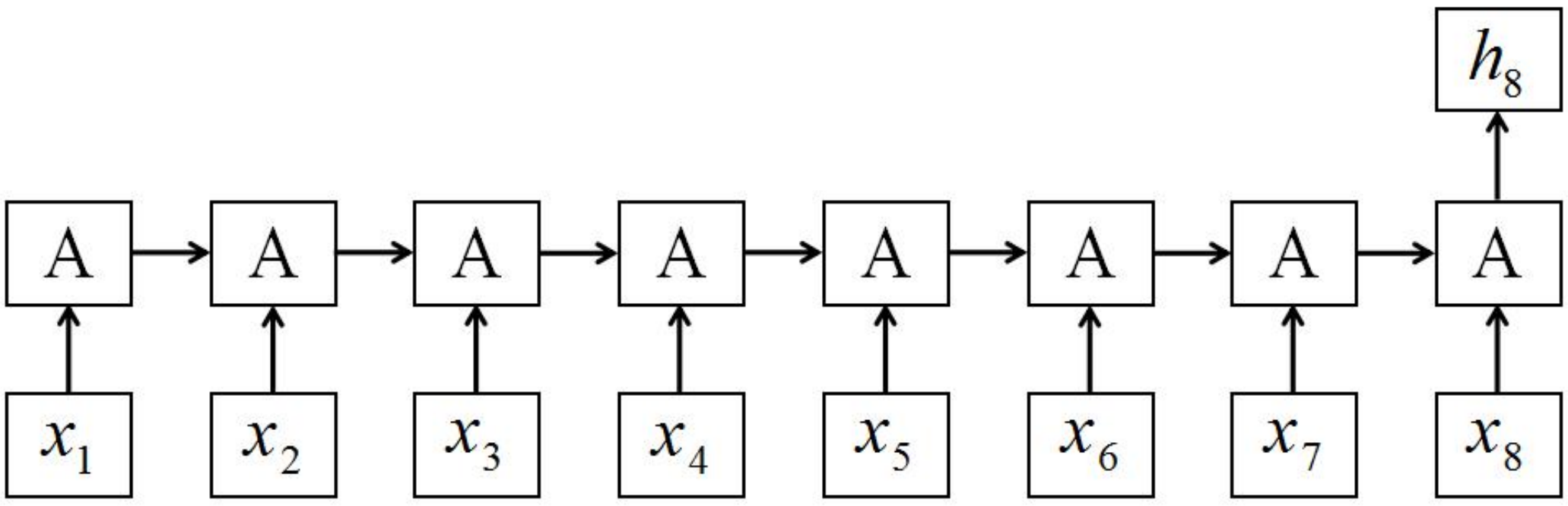

Figure 1: The standard RNN structure. Each step waits for the output of its previous step, which is computed by the recurrent unit A.

The length of the input sequence $X$ is $T$, and here we assume $T = 8$ as Figure 1 shown. The standard RNN uses the last hidden state as the representation of the whole sequence, and then add a softmax classifier to predict the labels. In addition to GRU and LSTM, QRNN and SRU could be seen as one form of recurrent unit A as well. However, the overall RNN structure has not been improved, at each step we need to wait for the output of the previous step:

$$h_t = f(h_{t-1}, x_t) \tag{5}$$

where $h$ is the previous hidden state. This standard RNN structure in which every two adjacent cells are connected causes the bottleneck: the longer the input sequence is, the longer it takes.

[1] https://github.com/zepingyu0512/srnn

## 2.3 Sliced Recurrent Neural Networks

We construct a new RNN structure called sliced recurrent neural networks (SRNNs), which is shown in Figure 2. In Figure 2 the recurrent unit is also referred to as A.

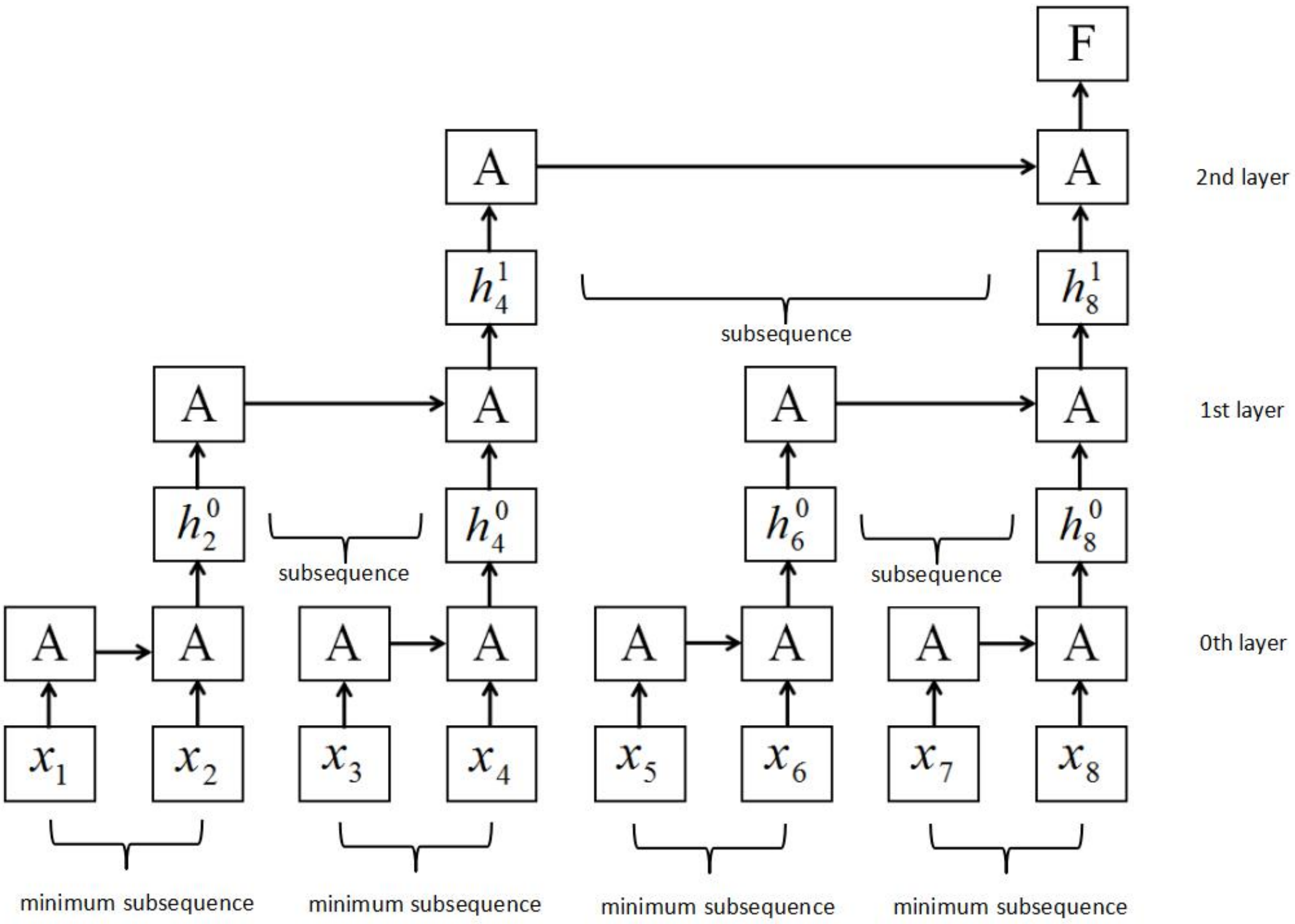

Figure 2: The SRNN structure. It is constructed by slicing the input sequence into several minimum subsequences with equal length. The recurrent units could work on each subsequence simultaneously on each layer, and the information could be transmitted through multiple layers.

The length of input sequence $X$ is $T$, and the input sequence is:

$$X = [x_1, x_2, ..., x_T] \tag{6}$$

where $x$ is the input at each step and it may have multiple dimensions such as word embeddings. Then we slice $X$ into $n$ subsequences of equal length, and the length of each subsequence $N$ is:

$$t = \frac{T}{n} \tag{7}$$

where $n$ is the slice number, and the sequence $X$ could be represented as:

$$X = [N_1, N_2, ..., N_n] \tag{8}$$

where each subsequence is:

$$N_p = [x_{(p-1)\times t+1}, x_{(p-1)\times t+2}, ..., x_{p\times t}] \tag{9}$$

Similarly, we slice each subsequence $N$ into $n$ subsequences of equal length again, and then repeat this slice operation $k$ times until we have an appropriate minimum subsequence length on the bottom layer (we call it 0th layer, which is shown in Figure 2), and $k$+1 layers are obtained by slicing $k$ times. The minimum subsequence length of 0th layer is:

$$l_0 = \frac{T}{n^k} \tag{10}$$

and the number of the minimum subsequences on 0th layer is:

$$s_0 = n^k \tag{11}$$

Because every parent sequence on pth layer (p>0) is sliced into $n$ parts, the number of the subsequences on pth layer is:

$$s_p = n^{k-p} \tag{12}$$

and the subsequence length of pth layer is:

$$l_p = n \tag{13}$$

Take Figure 2 as an example. The sequence length $T$ is 8, the slice operation times $k$ is 2, and the slice number $n$ of each pth layer is 2. After slicing the sequence twice, we get four minimum subsequences on 0th layer, and the length of each minimum subsequence is 2. If the length of the sequence or the length of its subsequences cannot be divided by $n$, we may exploit padding method or choose different slice number on each layer. Different $k$ and $n$ may be used on different tasks and datasets.

The difference between the SRNN and the standard RNN is that the SRNN slices the input sequence into many minimum subsequences and utilizes the recurrent units on each subsequence. In this way, the subsequences could be easily parallelized. On 0th layer, the recurrent units are acted on each minimum subsequence through the connection structure. Next, we obtain the last hidden states of each minimum subsequence on 0th layer, which are used as the input of their parent sequences on 1st layer. And then we use the last hidden state of each subsequence on (p-1)th layer as the input of their parent sequence on pth layer and compute the last hidden states of the subsequences on pth layer.

$$h_t^1 = \overrightarrow{GRU}^0(mss^0_{(t-l_0+1)\sim t}) \tag{14}$$

$$h_t^{p+1} = \overrightarrow{GRU}^p(h^p_{t-l_p} \sim h^p_t) \tag{15}$$

where $h_l^p$ is the number $l$ hidden state on pth layer, *mss* denotes minimum subsequences on 0th layer, and different GRUs could be used on different layers. This operation is repeated between each sub-parent sequence on each layer until we get the final hidden state $F$ of the top layer (kth layer).

$$F = \overrightarrow{GRU}^k(h^k_{t-l_k} \sim h^k_t) \tag{16}$$

### 2.4 Classification

Similar to the standard RNN, the softmax layer is added after the final hidden state $F$ to classify the labels:

$$p = \text{softmax}(W_{\text{F}}F + b_F) \tag{17}$$

and the loss function is negative log-likelihood:

$$loss = -\sum \log p_{dj} \tag{18}$$

where $d$ is each document of the dataset with label $j$.

### 2.5 Speed Advantage

The reason why SRNNs could be parallelized is that SRNNs improve the traditional connection structure. In SRNNs, not every current input is connected with its previous moment, but the entire sequence is connected together by a sliced method. SRNNs could also obtain the sequence order by the recurrent units in each subsequence, and transmit the information through multiple layers. We assume that the time spent in each recurrent unit is $r$, then the time spent in the standard RNN is:

$$t_{RNN} = T \times r \tag{19}$$

where $T$ is the input sequence length. In the SRNN, each minimum subsequence could be parallelized, so the time spent on 0th layer is:

$$t_0 = (\frac{T}{n^k}) \times r \tag{20}$$

and similarly, the time spent on pth layer (not including 0th layer) is:

$$t_p = n \times r \tag{21}$$

so the total time in the SRNN is:

$$t_{SRNN} = (n \times k + \frac{T}{n^k}) \times r \tag{22}$$

At last we could compute the speed advantage of the SRNN:

$$R = \frac{t_{SRNN}}{t_{RNN}} = \frac{1}{n^k} + \frac{n \times k}{T} \tag{23}$$

where R is how much faster the SRNN gets. We could choose different n and k to get different speed advantage.

### 2.6 Relations between the SRNN and the Standard RNN

In this part we describe the relations between the SRNN and the standard RNN. In the standard RNN structure, each step is related to the input and the previous step, which could be computed by:

$$h_t = f(Ux_t + Wh_{t-1} + b) \tag{24}$$

where $x$ is the input and $h$ is the hidden state. The function $f$ could be a nonlinear activation function such as sigmoid, or a linear activation function such as rectified linear units (Le et al., 2015). To simplify the question, here we discuss the case when we use a linear function:

$$f(x) = x \tag{25}$$

and we set the bias $b$ and $h_0$ to be zero. When we use the standard RNN, we could get the last hidden state:

$$h_T = Ux_T + Wh_{T-1} = Ux_T + W(Ux_{T-1} + Wh_{T-2}) = ... = Ux_T + WUx_{T-1} + W^2Ux_{T-2} + ... + W^{T-1}Ux_1 \tag{26}$$

where $T$ is the sequence length. And then we construct the SRNN structure. When $T = n^{k+1}$, we choose SRNN ($n$,$k$), which means slicing $k$ times with the slice number $n$. The SRNN has $k$+1 layers, on each layer the length of each subsequence is $n$. We could compute the last hidden state of each minimum subsequence on 0th layer:

$$h_n^0 = U_0x_n + W_0U_0x_{n-1} + W_0^2U_0x_{n-2} + ... + W_0^{n-1}U_0x_1$$

$$h_{2n}^0 = U_0x_{2n} + W_0U_0x_{2n-1} + W_0^2U_0x_{2n-2} + ... + W_0^{n-1}U_0x_{n+1}$$

$$...$$

$$h_T^0 = U_0x_T + W_0U_0x_{T-1} + W_0^2U_0x_{T-2} + ... + W_0^{n-1}U_0x_{T-n+1} \tag{27}$$

where $h_l^p$ is the number $l$ hidden state on pth layer. There are $n^k$ last hidden states on 0th layer. Similarly, we could take the hidden states on (p-1)th layer as the input, and compute the last hidden states of the subsequences on pth layer (p>0).

$$h_{n^{p+1}}^p = U_ph_{n^{p+1}}^{p-1} + W_pU_ph_{n^{p+1}-n^p}^{p-1} + W_p^2U_ph_{n^{p+1}-2n^p}^{p-1} + ... + W_p^{n-1}U_ph_{n^p}^{p-1}$$

$$h_{2n^{p+1}}^p = U_ph_{2n^{p+1}}^{p-1} + W_pU_ph_{2n^{p+1}-n^p}^{p-1} + W_p^2U_ph_{2n^{p+1}-2n^p}^{p-1} + ... + W_p^{n-1}U_ph_{n^{p+1}+n^p}^{p-1}$$

$$...$$

$$h_T^p = U_ph_T^{p-1} + W_pU_ph_{T-n^p}^{p-1} + W_p^2U_ph_{T-2n^p}^{p-1} + ... + W_p^{n-1}U_ph_{T-(n-1)n^p}^{p-1} \tag{28}$$

There are $n^{k-p}$ last hidden states on pth layer. And this operation is repeated from 0th layer to kth layer. At last we get the final hidden state $F$ of kth layer.

$$F = U_kh_T^{k-1} + W_kU_kh_{T-n^k}^{k-1} + W_k^2U_kh_{T-2n^k}^{k-1} + ... + W_k^{n-1}U_kh_{n^k}^{k-1} \tag{29}$$

When we compute equation (29) using each hidden state calculated by equation (27) and (28), we could get:

$$F = U_kU_{k-1}...U_0x_T + U_kU_{k-1}...W_0U_0x_{T-1} + U_kU_{k-1}...W_0^2U_0x_{T-2} + ... + W_k^{n-1}U_kW_{k-1}^{n-1}U_{k-1}...W_0^{n-1}U_0x_1 \tag{30}$$

When we compare equation (30) with equation (26), we could find that the two equations compute the same results when:

$$U_0 = U$$

$$U_k = U_{k-1} = \ldots = U_1 = I$$

$$W_k = W^{n^k} \quad (31)$$

where $I$ is the identity matrix, $U$ and $W$ are the parameters in equation (26). This means that SRNNs could have the same output as standard RNNs when equation (31) is satisfied. For example, when $T = 4$, $k = 1$ and $n = 2$, we get a SRNN (2,1) with two layers and compare it with the standard RNN. The last hidden state of the standard RNN is:

$$h_4 = Ux_4 + WUx_3 + W^2Ux_2 + W^3Ux_1$$

In SRNN (2,1), the hidden states on 0th layer are:

$$h_2^0 = U_0x_2 + W_0U_0x_1$$

$$h_4^0 = U_0x_4 + W_0U_0x_3$$

and the final hidden state of 1st layer is:

$$F = U_1h_4^0 + W_1U_1h_2^0 = U_1U_0x_4 + U_1W_0U_0x_3 + W_1U_1U_0x_2 + W_1U_1W_0U_0x_1$$

When we use equation (31) to set $W_0 = W$, $W_1 = W^2$, $U_0 = U$ and $U_1 = I$, we could get:

$$F = Ux_4 + WUx_3 + W^2Ux_2 + W^3Ux_1$$

which is equal to $h_4$ above.

We have proved that when the function $f$ is linear, the output of the SRNN is equal to the output of the standard RNN when the parameters satisfy equation (31), so the standard RNN is a special case of the SRNN. Furthermore, the SRNN may get high-level information when they have different parameters on different layers, so the SRNN is able to obtain more information from the input sequences than the standard RNN.

## 3 Experiments

### 3.1 Datasets

We evaluate SRNNs on six large-scale sentiment analysis datasets. Table 1 shows the information of the datasets. We choose 80% data for training, 10% for validation and 10% for testing.

| Dataset | Classes | Documents | Max words | Average words | Vocabulary |
|---|---|---|---|---|---|
| Yelp 2013 | 5 | 468,608 | 1060 | 129.2 | 202,058 |
| Yelp 2014 | 5 | 670,440 | 1053 | 116.1 | 210,353 |
| Yelp 2015 | 5 | 897,835 | 1092 | 108.3 | 228,715 |
| Yelp_P | 2 | 598,000 | 1073 | 139.7 | 308,028 |
| Amazon_F | 5 | 3,650,000 | 441 | 82.7 | 1,274,916 |
| Amazon_P | 2 | 4,000,000 | 257 | 80.9 | 1,348,126 |

Table 1: Dataset information. Max words denotes the max sequence length, and Average words denotes the average length of the sentences in each dataset.

**Yelp reviews**: The Yelp reviews datasets are obtained from the Yelp Dataset Challenge, which has 5 sentiment labels (the higher, the better). This dataset consists of 4,736,892 documents, and we extract three subsets Yelp 2013, 2014, 2015 containing 468,608, 670,440 and 897,835 documents separately. Zhang et al. (2015) created the polarity dataset including 598,000 documents with two sentiment labels, and we obtain the polarity dataset from them.

**Amazon reviews**: The Amazon reviews dataset is a commentary dataset containing 34,686,770 reviews on 2,441,053 products from 6,643,669 users (He and McAuley, 2016). One review has a review title, a review content and a sentiment label, and we combine the title and content into one document. The dataset is also constructed into a full dataset with 3,650,000 documents and a polarity dataset with 4,000,000 documents, which is also obtained from Zhang et al. (2015).

### 3.2 Baseline

We compare SRNNs with standard RNNs and take GRU as the recurrent unit. The output of the last hidden state is the representation of each document, and then we add a softmax layer on it to predict the sentiment labels. In order to compare SRNNs with convolutional structures, we also build a stack of dilated casual convolutional layers as a baseline, which is proposed in wavenet (Oord et al., 2016). The dilated casual convolutional structure could maintain the order information of the sequences. The dilation is 1, 2, 4, ... , 256 for Yelp datasets, and 1, 2, 4, ... , 128 for Amazon datasets. The number of filters is 50, and the activation function after each layer is rectified linear units.

### 3.3 Training

We use the sequence preprocessing tool of Keras (François et al, 2015) to pad sequences into the same length *T*. Sequences shorter than *T* are padded with zero at the end, and sequences longer than *T* are truncated. In this work, *T* is set to be 512 on Yelp datasets, and 256 on Amazon datasets. Different *n* and *k* values, which separately denotes the slice number and the slice times, are used on the experiments. For each dataset, we retain the top 30,000 words of the vocabulary. The pre-trained GloVe embeddings (Pennington et al., 2014) are utilized to initialize the word embeddings.

We set GRU as the recurrent unit of the SRNN. We have discussed the relations between the SRNN and the standard RNN in section 2.6 and have proved that the standard RNN is a special case of the SRNN when we use linear activation function between the recurrent units. However, it does not mean only linear activation function could be used in the SRNN. Both linear activation function such as hard sigmoid, and nonlinear activation function such as hyperbolic tangent could be used in or after each layer in the SRNN. In this paper, the recurrent activation function in GRU is sigmoid, and the activation function after each layer is linear function $f(x) = x$.

The dimension of GRU on each layer is set to be 50 and the word embedding dimension is 200. In SRNNs, the final state *F* also has 50 dimensions. We set the mini-batch size to be 100 for training and use Adam (Kingma and Ba, 2014) with $\alpha = 0.001$, $\beta_1 = 0.9$, $\beta_2 = 0.999$ and $\varepsilon = 10^{-8}$. We tune the hyper parameters on the validation set and select the best model to predict the sentiment labels on the test set. We train the models with an NVIDIA GTX 1080 GPU, and record the training time per epoch on each dataset.

### 3.4 Results and Analysis

The results on each dataset are shown on Table 2. We choose different *n* and *k* values and get different SRNNs. For example, SRNN (16,1) means *n*=16 and *k*=1, which could get a 32-length minimum subsequence when *T* is 512 or a 16-length minimum subsequence when *T* is 256. We compare four SRNNs with the standard RNN. For each dataset, we use bold words to label the highest-performing model and the fastest model.

| Dataset | Model | Parameters | Validation | Test | Time/Epoch |
|---|---|---|---|---|---|
| Yelp 2013 | GRU | 5.76M | 66.56 | 66.12 | 3172s |
| | SRNN (16,1) | 5.77M | 67.18 | **67.03** | 270s |
| | SRNN (8,2) | 5.79M | 67.11 | 66.80 | **145s** |
| | SRNN (4,3) | 5.80M | 67.26 | 66.72 | 164s |
| | SRNN (2,8) | 5.85M | 66.30 | 66.41 | 204s |
| | DCCNN | 5.78M | 64.91 | 64.79 | 67s |
| Yelp 2014 | GRU | 5.76M | 70.37 | 70.63 | 4142s |
| | SRNN (16,1) | 5.77M | 70.53 | 70.70 | 388s |
| | SRNN (8,2) | 5.79M | 70.35 | **70.76** | **201s** |
| | SRNN (4,3) | 5.80M | 70.25 | 70.48 | 238s |
| | SRNN (2,8) | 5.85M | 69.50 | 69.70 | 284s |
| | DCCNN | 5.78M | 68.46 | 68.66 | 96s |
| Yelp 2015 | GRU | 5.76M | 72.52 | 72.89 | 4434s |
| | SRNN (16,1) | 5.77M | 73.09 | **73.50** | 510s |
| | SRNN (8,2) | 5.79M | 72.84 | 73.30 | 319s |
| | SRNN (4,3) | 5.80M | 72.98 | 73.29 | **309s** |
| | SRNN (2,8) | 5.85M | 72.37 | 72.75 | 367s |
| | DCCNN | 5.78M | 70.69 | 70.94 | 131s |
| Yelp_P | GRU | 5.76M | 96.02 | 95.96 | 3170s |
| | SRNN (16,1) | 5.77M | 95.83 | 95.92 | 401s |
| | SRNN (8,2) | 5.79M | 95.87 | 95.99 | **205s** |
| | SRNN (4,3) | 5.80M | 95.90 | **96.04** | 236s |
| | SRNN (2,8) | 5.85M | 95.69 | 95.88 | 297s |
| | DCCNN | 5.78M | 95.03 | 95.26 | 98s |
| Amazon_F | GRU | 5.76M | 61.54 | 61.36 | 8953s |
| | SRNN (16,1) | 5.77M | 61.65 | **61.65** | 1584s |
| | SRNN (8,2) | 5.79M | 61.58 | 61.41 | **1147s** |
| | SRNN (4,3) | 5.80M | 61.50 | 61.40 | 1166s |
| | SRNN (2,7) | 5.85M | 61.04 | 60.88 | 1344s |
| | DCCNN | 5.78M | 59.64 | 59.60 | 401s |
| Amazon_P | GRU | 5.76M | 95.27 | 95.22 | 11062s |
| | SRNN (16,1) | 5.77M | 95.29 | **95.26** | 2144s |
| | SRNN (8,2) | 5.79M | 95.21 | 95.18 | **1309s** |
| | SRNN (4,3) | 5.80M | 95.12 | 95.12 | 1567s |
| | SRNN (2,7) | 5.85M | 94.98 | 95.02 | 1886s |
| | DCCNN | 5.78M | 94.72 | 94.69 | 448s |

Table 2: The accuracy and training time on validation and test sets of the models on each dataset. Four different structures of SRNNs are constructed. DCCNN is dilated casual convolutional neural network, which is described in section 3.2.

The results show that SRNNs achieve better performance and attain much faster speeds than standard RNNs on all the datasets with few extra parameters. Different SRNNs have achieved the best performance on different datasets. SRNN (16,1) gets the highest accuracy on Yelp 2013, Yelp 2015, Amazon_F and Amazon_P; SRNN (8,2) performs best on Yelp 2014; SRNN (4,3) is the best on Yelp_P. SRNNs with $k$ more than 1 could get nearly 15 times faster than standard RNNs on the Yelp datasets, and the speed advantage depends on $k$, $n$ and $T$. In this work, SRNN (4,3) has the fastest speed on Yelp 2015, while SRNN (8,2) is the fastest on the rest (except DCCNN).

When we focus on the results of SRNN (2,8) on Yelp datasets and SRNN (2,7) on Amazon datasets, we could find that even if they did not achieve the best performance, they did not lose much accuracy. This means that SRNNs are able to transmit information through multiple layers, and because of this, SRNNs may achieve remarkable results when we train very long sequences. When $n$ is 2, SRNN has the same number of layers as DCCNN, and it has much higher accuracy than DCCNN. So it means that the recurrent structure in SRNN is better than the dilated casual convolutional structure.

When we go back to equation (23), we may find that when $n$ and $k$ are not set to be too small, we could get much faster. In this work, we set $n=8$ , $k=q-1$ when $T=8^q$ . We use an NVIDIA GTX 1080 GPU to train the models on 5120 documents, since the standard RNN would take too much time if we use more data. The training time is shown on Table 3.

We could get the amazing results from Table 3: the longer the sequence length is, the bigger speed advantage the SRNN achieves. When the sequence length is 32768, SRNN would take only 52s while the standard RNN would take nearly 2 hours. The SRNN is 136 times as fast as the standard RNN, and the speed advantage may be even bigger when longer sequences are used. Therefore, SRNNs may achieve much faster speeds on long-sequence tasks such as speech recognition, character-level text classification and language modeling.

| Model | Sequence length | | |
| --- | --- | --- | --- |
| | $8^3 = 512$ | $8^4 = 4096$ | $8^5 = 32768$ |
| SRNN (8,2) | 4s | - | - |
| SRNN (8,3) | - | 9s | - |
| SRNN (8,4) | - | - | 52s |
| GRU | 54s | 476s | 7074s |
| Speed advantage | 13.5x | 52.9x | 136.0x |

Table 3: The training time and speed advantage on different sequence length. For each sequence length we choose a different SRNN structure.

### 3.5 Why SRNN

In this part we will discuss the advantages and importance of the SRNN. With the success of RNNs in many NLP tasks, many scholars have proposed different structures to increase the speed of RNNs. Most of the researches get faster by improving the recurrent units. However, the traditional connection structure has scarcely been questioned, in which each step is connected to its previous step. It is this connection structure that limits the speed of RNNs. The SRNN has improved the traditional connection method. Instead, a sliced structure is constructed to implement the parallelization of RNNs. The experimental results on six large-scale sentiment analysis datasets show that SRNNs achieve better performance than the standard RNN. The reasons are as follows:

(1) When we use the standard RNN connection structure, recurrent units with gated structures such as GRU and LSTM are useful, but they are not able to store all the important information when the sequences are very long. The SRNN, however, could divide the long sequence into many short subsequences and obtain the important information in short sequences. SRNNs are able to transmit the important information through the multiple-layers structure from 0th layer to the top layer.

(2) SRNNs have the ability to obtain high-level information from the sequences, instead of just the word-level information. When we use SRNN (8,2) in a document with 512 words, 0th layer may get the sentence-level information from the word embeddings, 1st layer may gain the paragraph-level information from the sentence-level information and 2nd layer could generate the final document-level representation from the paragraph-level information. The standard RNN, however, could only get the word-level information. Although it is impossible to have 8 paragraphs in each document, 8 sentences in each paragraph and 8 words in each sentence, the overall order information and structure information is uniform. Take the paragraph information as an example. People always express their opinions at the beginning or end of an article, and show examples in the middle of the article to explain their views. Compared to standard RNN, it is much easier for SRNNs to gain this information on the top layer.

(3) In terms of handling sequences, the SRNN is akin to the human brain mechanism. For example, if we, as humans, are given an article and asked to answer some questions about it, we usually do not need to read the whole article intensively. To answer the questions correctly, we try to locate the paragraph which mentions the specific information, and then find sentences and words in this paragraph that can answer the question. The SRNN could easily do this through multiple layers.

In addition to the improvement of accuracy, the most significant advantage of SRNNs is that SRNNs can be computed in parallel and achieve much faster speeds. Equation (23) computes the speed advantage of SRNN, and experiments of different sequence length also show that SRNNs could run much faster than standard RNNs. SRNNs could be even faster on longer sequences. As the Internet develops, hundreds of millions of data are generated every day, and SRNNs have devised new ways for us to handle these data.

## 4 Related Work

In order to increase the speed of RNNs, many scholars tried to improve the traditional RNN and achieved great results. Kim (2014), Kalchbrenner et al. (2014), and Gehring et al. (2017) tried to use CNNs in NLP tasks, which are usually used in computer vision (Lecun et al., 1998). Several structures get faster with the recurrent units changed (Greff et al., 2015; Balduzzi and Ghifary, 2016; Bradbury et al., 2017; Lei et al., 2017). As an overall structure improvement, the SRNN are related to these models, because different types of recurrent units could be used in SRNNs. In this work we choose GRU (Cho et al., 2014), but other recurrent units are able to work in SRNNs as well.

The SRNN is related to the hierarchical structure proposed by Tang et al. (2015) and Yang et al. (2016). Tang et al. (2015) use CNN or LSTM to obtain the sentence representations, and then exploit gated RNN to generate the document representations. Yang et al. (2016) build a hierarchical network on both word-level and sentence-level, then use attention mechanism on both level. The difference between the SRNN and the hierarchical structure is that the documents do not need to be split into sentences when we use the SRNN, and the SRNN could have multiple layers. The hierarchical structure could be viewed as a special case of the SRNN, where $k$ is 1 and all the sentences have equal length.

Several other architectures have been proposed to improve the connection structure of RNNs (Sutskever and Hinton, 2010; Koutnik et al., 2014; Chang et al., 2017). The SRNN is different from those architectures in the connection structure. The SRNN could be computed in parallel by slicing the sequences into many subsequences, but these models may still limit parallelization.

Also, the SRNN structure is similar to the overall structure of wavenet (Oord et al., 2016), which is used for audio generation. The difference between the SRNN and wavenet, which is also the most important innovation of the SRNN, is that we use recurrent units on each layer. For sequences, recurrent structure has its inherent advantages than convolutional structure.

## 5 Conclusion

In this paper we present the sliced recurrent neural network (SRNN), which is an overall structure improvement of RNN. The SRNN could reach a remarkably faster speed than the standard RNN and achieve better performance on six large-scale sentiment datasets.

## 6 Future Work

The SRNN has been successful in text classification. In future work, we hope to promote it to other NLP applications, such as question answering, text summarization, and machine translation. In sequence to sequence model, the SRNN can be used as the encoder, and the decoder may be improved by using an inverse SRNN structure. Also, we hope to use the SRNN in several long sequence tasks, such as language model, music generation and audio generation. And we want to explore more about variants of the SRNN. For example, bidirectional structure and attention mechanism could be added.

In section 2.6, we have discussed the relations between the SRNN and the standard RNN when choosing the linear activation function. In future work, we will try to research the situation of using nonlinear recurrent activation functions by mathematics, though it may be harder.

## Acknowledgements

Special thanks to Linjuan Wu for her help in polishing the language. Also, we would like to thank the anonymous reviewers for their suggestions. This work is supported by the National Natural Science Foundation of China (Grant No. 61772337, 61472248, and U1736207) and the SJTU-Shanghai Songheng Information Analysis Joint Lab.